\documentclass[letterpaper]{article}
\usepackage[preprint]{aaai2027}

\usepackage[hyphens]{url}
\usepackage{graphicx}
\usepackage{natbib}
\usepackage{caption}
\usepackage{amsmath}
\usepackage{amssymb}
\usepackage{booktabs}
\usepackage{colortbl}
\usepackage{multirow}
\usepackage{tikz}
\usetikzlibrary{positioning}

\definecolor{oursgray}{gray}{0.95}
\definecolor{bandgray}{gray}{0.92}

\newcommand{\method}{MoCRA}
\newcommand{\dataset}{UHV-4K-AIO}

\title{\method: Mixture of Compositional Rank-1 Atoms for 4K All-in-One Video Restoration}
\author{
    Yongcong Wang\textsuperscript{\rm 1}\thanks{ycwang1031@gmail.com},
    Pu Wang\textsuperscript{\rm 2,3},
    Hingchin Chen\textsuperscript{\rm 4},
    Runci Bai\textsuperscript{\rm 5},
    Yucheng Xin\textsuperscript{\rm 6},
    Chen Wu\textsuperscript{\rm 7},\\
    Chengchao Shen\textsuperscript{\rm 1},
    Guangwei Gao\textsuperscript{\rm 8},
    Siyuan Yao\textsuperscript{\rm 9},
    Pengwen Dai\textsuperscript{\rm 10},
    Zhuoran Zheng\textsuperscript{\rm 11}\thanks{Corresponding author: zhengzr@njust.edu.cn}
}
\affiliations{
    {\small
    \mbox{\textsuperscript{\rm 1}Central South University}\ 
    \mbox{\textsuperscript{\rm 2}Shandong University}\ 
    \mbox{\textsuperscript{\rm 3}Shenzhen Loop Area Institute}\ 
    \mbox{\textsuperscript{\rm 4}The Hong Kong University of Science and Technology}\ 
    \mbox{\textsuperscript{\rm 5}China Academy of Information and Communications Technology}\ 
    \mbox{\textsuperscript{\rm 6}Shandong Normal University}\ 
    \mbox{\textsuperscript{\rm 7}National University of Defense Technology}\ 
    \mbox{\textsuperscript{\rm 8}Nanjing University of Science and Technology}\ 
    \mbox{\textsuperscript{\rm 9}Beijing University of Posts and Telecommunications}\ 
    \mbox{\textsuperscript{\rm 10}Sun Yat-sen University}\ 
    \mbox{\textsuperscript{\rm 11}Independent Researcher}}
}

\begin{document}

\maketitle

\begin{abstract}
    Real-world video arrives hazy, rainy, dark, or noisy, and a deployable
    restorer faces three demands at once: no degradation label, native 4K
    output, and stability in playback. Existing methods answer them separately
    and break on the joint problem, because per-frame degradation readings flip
    between frames, downsampled proxies erase the rain and noise they are meant
    to remove, and dense temporal alignment does not fit 4K memory. No paired
    benchmark even poses that problem, so we build one. UHV-4K-AIO renders
    physically modeled haze, rain, sensor noise, and low light over the same
    100 clean 4K clips with shared depth and motion, and its construction
    exposes the split MoCRA is built on: haze and low light survive aggressive
    downsampling, while rain and noise exist only at native scale. Band-matched
    compositional conditioning follows, spending conditioning capacity,
    computation, and supervision in the band where each degradation lives. One
    dictionary of rank-1 atoms, recomposed sparsely per frame, conditions both
    a once-per-clip coarse branch and a shallow native-resolution refiner, in
    3.6M parameters and with no optical flow. Trained once for all four tasks,
    MoCRA takes the best task-mean PSNR of eleven retrained image and video
    baselines, holds warping error at the level of the flow-based video models
    while never estimating motion, and restores native 4K in under half a
    second, against 1.7 seconds for the fastest baseline.
    \end{abstract}

\section{Introduction}

\begin{figure}[t]
\centering
\includegraphics{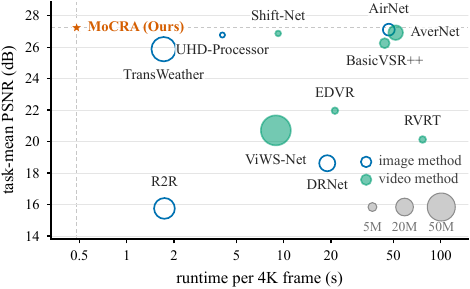}
\caption{\textbf{Task-mean PSNR against per-frame 4K runtime};
bubble area is proportional to parameter count. No baseline is both faster
and more accurate than \method{}.}

\label{fig:teaser}
\end{figure}

Video captured outside controlled conditions is degraded by its own
acquisition chain: haze from atmospheric scattering, rain streaks, a starved
exposure in low light, and the noise amplified by the gain that
compensates for it. A restorer deployed against this
chain is not told which degradation it faces, must produce the
$3840\times2160$ frames the camera recorded, and is judged in playback, where
errors too small to move a frame metric remain visible as flicker. Meeting
these three demands one at a time is expensive on our benchmark
(Figure~\ref{fig:teaser}): the accurate methods take tens of seconds per 4K
frame, the fast ones give up fidelity.

Each demand has a mature answer, and each answer rests on an assumption the
joint problem breaks. All-in-one image restoration reads the degradation off
a single image as a prompt or an embedding \citep{airnet,promptir,adair};
repeated per frame, the reading can flip between neighbors, and the flip
lands in the output as a color step. Ultra-high-definition restoration
estimates its correction on a downsampled proxy
\citep{dehaze4kid,uhdprocessor} that retains smooth haze and low light but
has already erased the rain streaks and sensor noise it is asked to remove.
Video restoration buys fidelity with per-pixel flow and dense temporal
attention \citep{basicvsrpp,rvrt}, whose memory grows with the pixel count,
so training stays on small crops and evaluation below 1080p. A pipeline that
stacks a prompt, a proxy, and a flow module inherits all three failures at
once.

A harder obstacle sits underneath: there is no data on which the joint
problem can be posed. The same 4K scene cannot be captured hazy, rainy,
dark, noisy and clean, and benchmarks pooled from a different single-task
source per task change the content along with the degradation, so task
interference cannot be told apart from content shift. \dataset{} closes that
gap: four physically modeled degradations over one shared set of clean 4K
clips, so the tasks differ in nothing but the degradation.

The benchmark's own construction points to the model. Haze and low light are
smooth fields whose evidence survives heavy downsampling; rain streaks and
sensor noise live in individual native pixels that no downsampled view can
carry. \method{} turns this split into a design principle, band-matched
compositional conditioning: conditioning capacity, computation, and
supervision each go to the band where a degradation lives.

At 3.6M parameters, \method{} leads eleven baselines retrained under the same
recipe on task-mean PSNR, at half a second and 3.9\,GiB per 4K frame. This
paper contributes:
\begin{itemize}
\item \textbf{A setting that makes the problem testable.} \dataset{} pairs
four degradations as native 4K video over one shared set of clean frames,
scored on fidelity, no-reference quality, temporal stability and cost.
\item \textbf{A model that spends where the degradation is.} \method{}
realizes the principle with one routing space across both bands and
supervision matched to the same split.
\item \textbf{Evidence across four metric families.} \method{} matches the
steadiest baselines in warping error without estimating motion, holds its
lead when the degradation switches inside a clip, and wins on real captured
haze; ablations tie each margin to one allocation.
\end{itemize}

\section{Related Work}

\noindent\textbf{All-in-one and video restoration.}
One model covers several corruptions by reading the degradation off the image
and injecting it back as conditioning: contrastive embeddings and learnable
prompts \citep{airnet,promptir}, frequency-band modulation \citep{adair}, or
mixtures of low-rank adapters \citep{hu2022lora,lorair,uirlora}. In each case
the degradation is a static attribute of one low-resolution image and the
composed unit is a whole prompt or adapter at a single scale. Video restoration
instead aligns and fuses neighbors, and pays for it: flow-guided propagation
\citep{basicvsrpp} and temporal mutual attention \citep{vrt,rvrt} hold training
crops to a few hundred pixels per side and evaluation at 1080p and below, with
separate weights per task.
Blind video models remove weather without a label \citep{viwsnet}, realign
frames as the corruption switches \citep{avernet}, or carry a causal latent
history across tasks \citep{turtle}, but none pairs the four acquisition-side
degradations we target. \method{} keeps the low-rank vocabulary but learns per-site rank-1
atoms from scratch, composes them sparsely per frame, and threads one routing
space through both scales.

\noindent\textbf{Ultra-high-definition restoration.}
The cost of computing at native resolution forces a low-resolution proxy,
from which the correction is transferred upward: bilateral grids sliced at 4K
\citep{dehaze4kid}, Fourier-amplitude correction on a downsampled image
\citep{uhdll}, or a compact latent shared by several UHD tasks
\citep{uhdprocessor}. The proxy holds for haze and illumination but breaks for
rain streaks and sensor noise, and estimating the global parameters per frame
leaves a video extension with color drift and flicker. \method{} keeps the
proxy only for the bands that survive it and gives rain and noise a
native-scale path of their own.

\section{UHV-4K-AIO Dataset}

\begin{figure}[t]
\centering
\includegraphics[width=\linewidth]{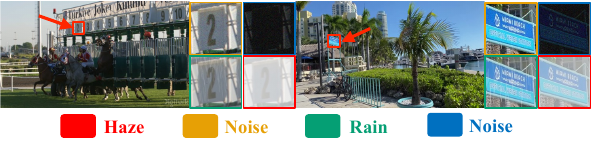}
\vspace{-18pt}
\caption{\textbf{Samples of \dataset{}.} Two clips, each with clean
frame and native-resolution crops of four degradations.}
\vspace{-10pt}
\label{fig:dataset}
\end{figure}

Paired restoration data is all-in-one only at 1080p and below
\citep{viwsnet,avernet}, 4K for one degradation at a time
\citep{uhdvd,realisvsr,libranet}, or real captured pairs below 4K
\citep{revide,sdsd}; Table~\ref{tab:datasets-full} places \dataset{} among them.
Upsampling lifts none of it: neither the one-pixel geometry of rain streaks
nor the per-pixel statistics of noise survives interpolation.
\dataset{} therefore extends UHV-4K \citep{libranet}, a paired 4K video
dehazing dataset, from one degradation to four. Its 100 clean clips and
their cached depth, transmission and flow carry over, and rain, noise and
low light are synthesized beside haze on that substrate
(Figure~\ref{fig:dataset}). The inherited 80/20 clip split holds the same 20
clips out for every task: 1{,}680 training and 420 test five-frame windows
each.

\subsection{Degradation Synthesis}
Every degradation comes from an explicit imaging or sensor model whose
parameters are frozen per clip and graded into three severity tiers, as a
single capture would fix them. Depth and flow are estimated once per clip
on a low-resolution proxy and shared by all four branches; smooth fields are
computed there and upsampled, while rain streaks are rasterized at 4K, where
upsampling would attenuate them. Haze is adopted unchanged from UHV-4K
\citep{libranet}: atmospheric scattering at a clip-level coefficient and
atmospheric light, with depth smoothed along the flow so the veil holds still
under camera motion. The other three are new, and in the fine band temporal
correctness is not automatic.

\noindent\textbf{Rain.}
Rain composites a streak layer over a thin veil,
$I_{\mathrm{rain}}=(1-m)\,I_{\mathrm{veil}}+m\,L_s$, the opacity mask $m$
rasterized at 4K and the veil scattered at $\beta_r\!\ll\!\beta$ so rain and
haze stay separable. Temporal coherence comes from propagating a particle
field rather than warping a rendered streak image, which would smear the
structures deraining must remove:
\begin{equation}
p_t = p_{t-1} + \mathrm{flow}(p_{t-1}) + v\,(\sin\theta,\ \cos\theta),
\end{equation}
at tier-level fall speed $v$ and clip-level angle $\theta$, following
RDD-Net \citep{rddnet}.

\noindent\textbf{Noise.}
Noise is a camera sensor response in the linear domain,
\begin{equation}
\begin{gathered}
I_{\text{noise}} = \Phi\!\big(g_t\,x_t + n^{\text{sr}}_t
+ n^{\text{fpn}} + n^{\text{band}}_t\big),\\[1pt]
n^{\text{sr}}_t \sim \mathcal{N}\big(0,\ \sigma_{\mathrm{s}}^{2}g_t\,x_t
+ \sigma_{\mathrm{r}}^{2}\big),
\end{gathered}
\end{equation}
with $\Phi$ the return to 8-bit sRGB and the signal-dependent variance the
heteroscedastic model of \citet{foi2008}, calibrated on real sensors
\citep{crvd}. Its terms move in time as a sensor's do: the fixed pattern
$n^{\text{fpn}}$ stays put in image coordinates while the scene moves
\citep{starlight}, shot-read noise and the row-column banding
$n^{\text{band}}_t$ are redrawn every frame, and the gain $g_t$ drifts as a
slow AR(1) exposure flicker.

\noindent\textbf{Low light.}
Low light cascades an exposure drop $a\,x_t^{\gamma}$ with $\gamma>1$, a
clip-level per-channel color cast $\mathbf{g}$ and a radial vignette $V(r)$,
then adds the same sensor noise. Darker tiers are tied to a higher ISO, so
the noise grows with the darkening \citep{lin2025lowlight,sdsd}, and the
released illumination map $a\,\mathbf{g}\,V(r)$ carries neither gamma nor
noise.

\dataset{} also releases the physical state behind each frame: depth and
transmission, the shared flow, the per-pixel noise-level map, and the
illumination map above, none of which other all-in-one benchmarks ship. The
flow doubles as the reference motion for the temporal metrics below. Tier
settings, constants, seeding, and a consistency protocol that re-estimates each
assigned parameter out of the released frames are deferred to the appendix.

\section{Methodology}

\begin{figure*}[t]
\centering
\includegraphics[width=0.87\textwidth]{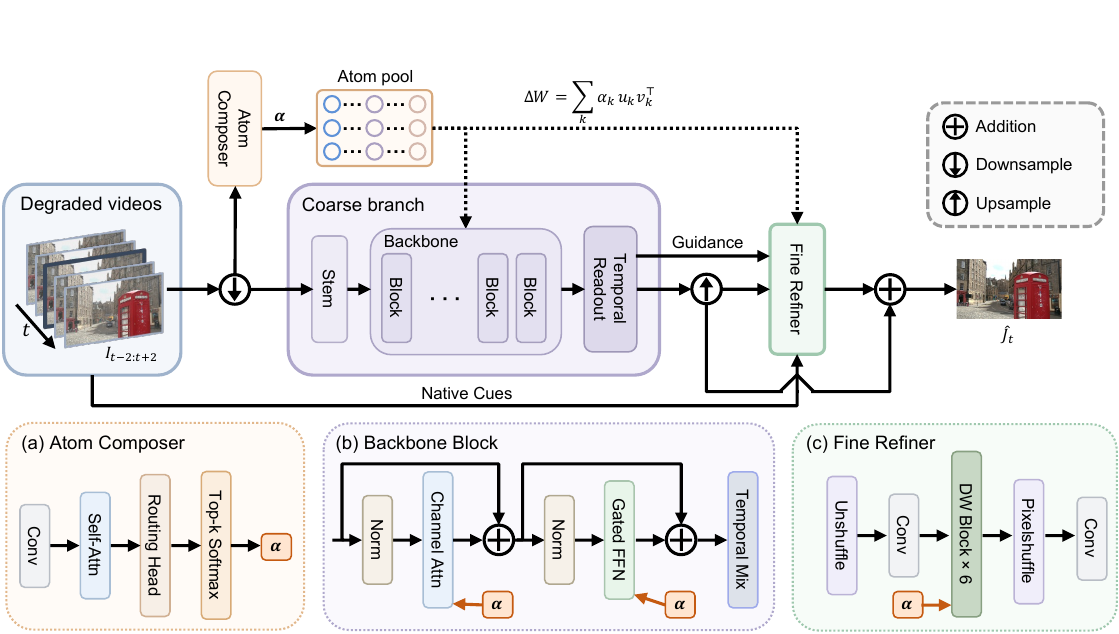}
\vspace{-4pt}
\caption{\textbf{Overview of \method{}.} The window is reduced once to
$I^{\downarrow}$, where the Atom Composer emits $\alpha$ and the coarse branch
restores global photometry; only the Fine Refiner reads native pixels. Dashed
arrows carry $\alpha$. Insets: (a) composer, (b) backbone block, (c) refiner.}
\vspace{-10pt}
\label{fig:mocra}
\end{figure*}

Given five consecutive degraded frames, \method{} predicts the clean center
frame $J_t$ as $\hat{J}_t=\mathcal{M}_\theta(I_{t-2},\ldots,I_{t+2})$, with
one set of weights for the four degradations of \dataset{} and no
degradation label at inference (Figure~\ref{fig:mocra}). The superscript
$\downarrow$ marks the aspect-preserving reduction of a frame to a short side
of 512, an antialiased $4.2\times$ decimation that erases structure finer
than roughly eight native pixels: it is the operational form of the band
split \method{} is organized around, since haze and low light stay legible in
$I^{\downarrow}$ while rain streaks and per-pixel noise do not.

\subsection{Rank-1 Atoms and Per-Frame Routing}
An all-in-one model must decide what to share across degradations and what to
specialize, and both fixed answers fail: a shared network compromises between
corrections that pull in opposite directions, and one expert per degradation
hard-wires a closed list that unknown or mixed input breaks. \method{} makes
the partition learnable, decomposing behavior into small reusable units and
specializing only their mixture.

\noindent\textbf{Per-site dictionaries.}
An \emph{injection site} is a linear map $W_l$ whose weight \method{} adapts
per frame. Each of the $L{=}68$ sites carries a dictionary of $K{=}96$ rank-1
atoms, factor pairs $u_{l,k}\in\mathbb{R}^{d_l^{\mathrm{out}}}$ and
$v_{l,k}\in\mathbb{R}^{d_l^{\mathrm{in}}}$ matched to its width and common to
all four degradations; fifty-six sit in the backbone's attention and
feed-forward projections, twelve in the pointwise convolutions of the
refiner. Rank one is the smallest unit of behavior a weight update can share,
and cheap enough at $K(d_l^{\mathrm{out}}{+}d_l^{\mathrm{in}})$ parameters to
condition every site of both branches.

\noindent\textbf{Frame-conditioned low-rank adaptation.}
At site $l$ and frame $t$ a sparse coefficient vector
$\alpha_{t,l}\in\mathbb{R}^{K}$ with $r{=}12$ active entries mixes the atoms
into a weight update that the site's activation $h$ passes through,
\begin{equation}
\Delta W_l(t)=\sum_{k=1}^{K}\alpha_{t,l,k}\,u_{l,k}v_{l,k}^{\top},
\qquad h\mapsto\bigl(W_l+\Delta W_l(t)\bigr)h,
\label{eq:atoms}
\end{equation}
so sites share the composition rule and one routing space, never a single
matrix. The mixture conditions the operator rather than the activation: a
prompt changes what a fixed layer sees, $\Delta W_l(t)$ changes what it does.

\noindent\textbf{The Atom Composer.}
One Atom Composer produces the whole routing tensor in a single pass over the
clip, reading $I^{\downarrow}$ reduced further to a short side of 256.
Degradation evidence is spatially uneven, haze thickens with depth, so global
pooling would collapse its layout to a mean. The composer instead appends
four learnable register tokens \citep{darcet2024registers} to the spatial
tokens of a small convolutional stem, mixes them in one self-attention layer,
and reads back only the registers. It never predicts a degradation class: a
factorized head maps the registers straight to routing logits
$z\in\mathbb{R}^{T\times L\times K}$ over the $T{=}5$ frames.

\noindent\textbf{Sparse routing.}
Temperature-scaled top-$r$ selection turns the logits into the coefficients
of Eq.~\ref{eq:atoms}, with exploration noise during training and none at
inference: writing $\tilde{z}_{t,l,k}=z_{t,l,k}+\xi_{t,l,k}$,
$\xi_{t,l,k}\sim\mathcal{N}(0,\sigma^2)$, and
$\mathcal{A}_{t,l}=\operatorname{Top}_r(\tilde{z}_{t,l,\cdot})$,
\begin{equation}
\alpha_{t,l,k}=
\begin{cases}
\dfrac{\exp(\tilde{z}_{t,l,k}/\tau)}
{\sum_{j\in\mathcal{A}_{t,l}}\exp(\tilde{z}_{t,l,j}/\tau)},
& k\in\mathcal{A}_{t,l},\\[6pt]
0, & \text{otherwise},
\end{cases}
\end{equation}
in the sparsely gated tradition of \citet{shazeer2017moe}. Routing
specializes only over atoms that already mean something, and atoms
differentiate only under routing that commits; annealing $\tau$ from 5 to 0.4
and $\sigma$ from 0.3 to zero over the first 15k steps breaks that circle.
Backbone sites read their own frame's routing, refiner sites the center
frame's. This tensor is the model's entire degradation judgment, so jitter
between frames would reach the output as a color step; a penalty
$\mathcal{L}_{t\alpha}=\operatorname{mean}\bigl((\alpha_{t}-\alpha_{t-1})^{2}\bigr)$
suppresses it at the source.

\subsection{Dual-Band Execution at Native 4K}
\noindent\textbf{Coarse branch.}
Degradation analysis, routing and coarse photometric restoration all run on
the five-frame $I^{\downarrow}$, at a cost independent of the output
resolution. The backbone pairs channel attention, linear in area, with a gated
depthwise feed-forward \citep{restormer} over fourteen blocks, and a depthwise
convolution along the frame axis couples the window. Each frame's latent
decodes to a coarse restoration $Y_t$; for the center frame a Temporal
Consensus Readout fuses the window by a softmax over learned per-position,
per-frame scores, so static regions average over the window and moving ones
follow the center frame. No motion field is affordable: an
all-pairs correlation volume at one eighth of 4K already needs about
63\,GiB in single precision. Because this branch sets the global photometry of
every frame, a gated term holds it steady wherever the degraded input is
static,
\begin{equation}
\begin{gathered}
\mathcal{L}_{tY}=\operatorname{mean}\bigl(G_t\odot|Y_t-Y_{t-1}|\bigr),\\[1pt]
G_t=\exp\Bigl(-\operatorname{box}_7\bigl(
\operatorname{mean}_c|I^{\downarrow}_t-I^{\downarrow}_{t-1}|\bigr)/\rho\Bigr),
\end{gathered}
\end{equation}
with $\operatorname{box}_k$ a $k\times k$ box mean and $\rho=0.05$.

\noindent\textbf{Native-scale refiner.}
What the coarse branch leaves behind is local, and one shallow module handles
it. The refiner takes the upsampled coarse decoding $\bar{Y}_t$ as its base
and adds cues that separate the high frequencies to remove from those to
keep: the degraded center frame with its Sobel magnitude, and the two
adjacent frames with their absolute differences. Transience marks the
degradation itself: a rain streak almost never repeats across frames,
inter-frame high-frequency correlation being 0.01 on our data, and noise is
redrawn at every pixel, whereas texture persists. Six
conditioned blocks at half resolution, receptive field near forty native
pixels, then predict a high-frequency residual over that base.

\noindent\textbf{Cross-scale conditioning.}
The refiner never judges the degradation on its own; the coarse branch
conditions it three ways. Its twelve atom sites bring the rank-1 mixture into
the blocks. Feature-wise modulation
\citep{perez2018film} pools $\alpha$ over sites and joins it with a Sobel
summary of the native window, so texture the coarse view cannot resolve still
reaches the conditioning. Cross-scale guidance projects the center-frame
backbone latent onto the refiner's grid, adding spatial layout a routing
vector cannot carry.

\noindent\textbf{Inference at 4K.}
The routing tensor, coarse output and guidance latent are computed
once per window, never inside it; the refiner runs on the full frame and falls
back to overlapping cosine-tapered tiles only when the frame does not fit in
memory. Tiling then touches a local operator alone, which removes the
color drift of tiled global estimation. Training replays the path, so the
coarse branch is supervised at test scale; the appendix gives
the full procedure.

\subsection{Band-Matched Supervision}
The fine band is also where a uniform loss fails. Rain streaks cover about
3\% of a rainy frame and take about 6\% of the gradient of a uniform pixel
loss, leaving the refiner a cheaper strategy than removal: copy native high
frequencies onto the coarse base, a shortcut that frame-level PSNR barely
punishes. Under
that loss our refiner passes 0.81 of the high-frequency energy inside
rain-streak masks through to the output, against 0.16 for noise with the same
weights. Capacity is not the bottleneck; the loss is spent where the
degradation is not.

\begin{figure}[tb]
\centering
\includegraphics{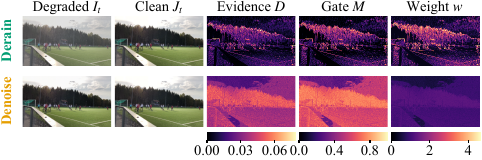}
\vspace{-6pt}
\caption{\textbf{Degradation-evidence weighting} on one held-out clip rendered
under rain and under noise. Rain concentrates $w$ on the structure it altered;
noise leaves it near uniform.}
\vspace{-10pt}
\label{fig:dew}
\end{figure}

\noindent\textbf{Degradation-evidence weighting (DEW).}
The remedy moves fine-scale supervision onto the degradation's own footprint;
Figure~\ref{fig:dew} traces the construction of its map $w$. With
$\mathrm{HP}(f)=f-\operatorname{box}_{17}(f)$ the complementary high-pass and
$\operatorname{mean}_c$ the average over color channels,
\begin{equation}
\begin{gathered}
D=\operatorname{box}_3\bigl(\operatorname{mean}_c
\bigl|\mathrm{HP}(I_t-J_t)\bigr|\bigr),\\[1pt]
M=\frac{D^2}{D^2+s^2},
\qquad
w=\frac{1+\lambda M}{\operatorname{mean}(1+\lambda M)},
\end{gathered}
\label{eq:dew}
\end{equation}
with $\lambda=24$, $s=0.03$, and the normalizing mean taken per sample;
$\lambda=0$ recovers the uniform loss exactly.

The evidence is degraded minus clean, so $w$ is static and decoupled from
prediction error, unlike focal-style weights that chase whatever the model
currently gets wrong \citep{jiang2021focal}, and the high-pass selects exactly
the structure $I^{\downarrow}$ cannot represent. The saturating gate
\citep{gemanmcclure1987} reports presence rather than magnitude. DEW is
training protocol, not architecture: it costs nothing at inference and every
model we train gets it, baselines included.

\begin{table*}[t]
    \centering
    \caption{\textbf{Comparison on \dataset{} at native 4K}, every method
    retrained under one recipe. No-reference scores average the four tasks.
    The gray row scores the degraded input; bold and underline mark best and
    second best among all video methods.}
    \label{tab:main}
    \scriptsize
    \setlength{\tabcolsep}{2.5pt}
    \resizebox{\textwidth}{!}{%
    \begin{tabular}{ll|ccc|ccc|ccc|ccc|ccc|cc|cc}
    \toprule
    \multirow{2}{*}{Method} & \multirow{2}{*}{Venue}
    & \multicolumn{3}{c|}{Dehaze} & \multicolumn{3}{c|}{Derain}
    & \multicolumn{3}{c|}{Denoise} & \multicolumn{3}{c|}{Low light}
    & \multicolumn{3}{c|}{Mean} & \multicolumn{2}{c|}{No-reference}
    & \multicolumn{2}{c}{Efficiency at 4K} \\
    \cmidrule(lr){3-5}\cmidrule(lr){6-8}\cmidrule(lr){9-11}
    \cmidrule(lr){12-14}\cmidrule(lr){15-17}\cmidrule(lr){18-19}\cmidrule(l){20-21}
    & & PSNR$\uparrow$ & SSIM$\uparrow$ & LPIPS$\downarrow$
    & PSNR$\uparrow$ & SSIM$\uparrow$ & LPIPS$\downarrow$
    & PSNR$\uparrow$ & SSIM$\uparrow$ & LPIPS$\downarrow$
    & PSNR$\uparrow$ & SSIM$\uparrow$ & LPIPS$\downarrow$
    & PSNR$\uparrow$ & SSIM$\uparrow$ & LPIPS$\downarrow$
    & MUSIQ$\uparrow$ & MANIQA$\uparrow$
    & Params\,(M) & Runtime\,(s) \\
    \midrule
    \textcolor{black!55}{Degraded input} & \textcolor{black!55}{--} & \textcolor{black!55}{13.19} & \textcolor{black!55}{0.769} & \textcolor{black!55}{0.226} & \textcolor{black!55}{27.24} & \textcolor{black!55}{0.916} & \textcolor{black!55}{0.131} & \textcolor{black!55}{27.73} & \textcolor{black!55}{0.558} & \textcolor{black!55}{0.493} & \textcolor{black!55}{9.06} & \textcolor{black!55}{0.232} & \textcolor{black!55}{0.966} & \textcolor{black!55}{19.31} & \textcolor{black!55}{0.619} & \textcolor{black!55}{0.454} & \textcolor{black!55}{31.14} & \textcolor{black!55}{0.223} & \textcolor{black!55}{--} & \textcolor{black!55}{--} \\
    \midrule
    \rowcolor{bandgray}\multicolumn{21}{c}{\emph{(a) Image methods}} \\
    \midrule
    AirNet & CVPR'22 & 22.04 & 0.939 & 0.051 & 28.99 & 0.973 & 0.022 & 37.54 & 0.968 & 0.036 & 19.86 & 0.826 & 0.221 & 27.11 & 0.927 & 0.083 & 36.04 & 0.230 & 8.93 & 46.69$^\dagger$ \\
    TransWeather & CVPR'22 & 19.39 & 0.883 & 0.105 & 27.94 & 0.918 & 0.129 & 35.15 & 0.940 & 0.110 & 20.98 & 0.765 & 0.347 & 25.87 & 0.877 & 0.173 & 33.50 & 0.222 & 38.05 & 1.72$^\dagger$ \\
    UHD-Processor & CVPR'25 & 20.38 & 0.899 & 0.106 & 29.79 & 0.934 & 0.113 & 33.09 & 0.784 & 0.340 & 23.82 & 0.693 & 0.490 & 26.77 & 0.828 & 0.262 & 32.60 & 0.210 & 1.71 & 4.07$^\dagger$ \\
    DRNet & TMM'26 & 14.16 & 0.812 & 0.183 & 26.34 & 0.916 & 0.127 & 24.29 & 0.806 & 0.270 & 9.67 & 0.369 & 0.686 & 18.62 & 0.726 & 0.317 & 33.06 & 0.212 & 16.49 & 18.93$^\dagger$ \\
    R2R & CVPR'26 & 11.74 & 0.752 & 0.264 & 16.49 & 0.854 & 0.163 & 25.68 & 0.880 & 0.177 & 9.08 & 0.332 & 0.676 & 15.75 & 0.705 & 0.320 & 31.34 & 0.219 & 27.58 & 1.74$^\dagger$ \\
    \midrule
    \rowcolor{bandgray}\multicolumn{21}{c}{\emph{(b) Video methods}} \\
    \midrule
    EDVR & CVPRW'19 & 21.08 & 0.921 & 0.093 & 23.40 & 0.937 & 0.081 & 24.11 & 0.636 & 0.272 & 19.25 & 0.533 & 0.412 & 21.96 & 0.757 & 0.214 & 34.12 & 0.212 & 2.85 & 21.19$^\dagger$ \\
    BasicVSR++ & CVPR'22 & 21.78 & \textbf{0.932} & 0.061 & 26.31 & \underline{0.955} & \underline{0.059} & 34.55 & \underline{0.948} & \underline{0.062} & \underline{22.34} & 0.819 & \underline{0.246} & 26.25 & \underline{0.914} & \underline{0.107} & \underline{35.81} & 0.218 & 6.45 & 43.93$^\dagger$ \\
    RVRT & NeurIPS'22 & 19.77 & 0.827 & 0.191 & 23.63 & 0.899 & 0.148 & 19.83 & 0.842 & 0.189 & 17.29 & 0.671 & 0.449 & 20.13 & 0.810 & 0.244 & 35.79 & 0.212 & 3.06 & 76.62$^\dagger$ \\
    Shift-Net & CVPR'23 & \underline{22.24} & \underline{0.924} & \underline{0.054} & 27.28 & 0.919 & 0.128 & \underline{36.19} & 0.917 & 0.142 & 21.76 & 0.765 & 0.352 & 26.87 & 0.881 & 0.169 & 34.22 & 0.219 & 2.17 & 9.22$^\dagger$ \\
    ViWS-Net & ICCV'23 & 22.23 & 0.913 & 0.068 & 23.59 & 0.907 & 0.096 & 28.57 & 0.894 & 0.154 & 8.44 & 0.279 & 0.573 & 20.71 & 0.748 & 0.223 & 33.08 & 0.221 & 57.67 & \underline{8.91}$^\dagger$ \\
    AverNet & NeurIPS'24 & 21.25 & 0.922 & 0.063 & \underline{27.58} & \textbf{0.957} & \textbf{0.039} & \textbf{37.32} & \textbf{0.957} & \textbf{0.050} & 21.54 & \underline{0.830} & \textbf{0.222} & \underline{26.92} & \textbf{0.917} & \textbf{0.094} & \textbf{36.07} & \textbf{0.225} & 13.72 & 51.63$^\dagger$ \\
    \midrule
    \rowcolor{oursgray}\textbf{\method{} (Ours)} & -- & \textbf{23.47} & \textbf{0.932} & \textbf{0.050} & \textbf{28.70} & 0.931 & 0.095 & 32.07 & 0.907 & 0.165 & \textbf{24.66} & \textbf{0.834} & 0.252 & \textbf{27.23} & 0.901 & 0.141 & 34.40 & \underline{0.224} & 3.64 & \textbf{0.48} \\
    \bottomrule
    \end{tabular}}
    \vspace{-4pt}
    \end{table*}

\noindent\textbf{Training objective.}
Each reconstruction term supervises the band it belongs to. Carrying the
evidence map, $\mathcal{L}^{w}_{\mathrm{pix}}=\operatorname{mean}\bigl(w
\odot\mathrm{Ch}(\hat{J}_t,J_t)\bigr)$ uses the Charbonnier penalty
$\mathrm{Ch}(a,b)=\sqrt{(a-b)^2+\varepsilon^2}$ with $\varepsilon=10^{-3}$
\citep{charbonnier1994}, and $\mathcal{L}^{w}_{\mathrm{grad}}$ applies the
same weighted penalty to Sobel magnitudes, setting the fine band's absolute
budget. The other two do not concern that band and stay
unweighted: $\mathcal{L}_{\mathrm{coarse}}=\operatorname{mean}\,
\mathrm{Ch}(\bar{Y}_t,J^{\downarrow}_t)$ supervises the coarse output at its
test-time scale, and $\mathcal{L}_{\mathrm{perc}}
=\tfrac{1}{4}\sum_i\operatorname{mean}\bigl|\phi_i(\hat{J}_t)
-\phi_i(J_t)\bigr|$ compares features $\phi_i$ from four blocks of a frozen
DINOv2 \citep{dinov2}. An incoherence penalty
$\mathcal{L}_{\mathrm{inc}}$ on overlapping atom pairs and a load-balance term
$\mathcal{L}_{\mathrm{bal}}$ against dead atoms guard the dictionary, both
defined in the appendix. With $n$ the optimizer step,
\begin{equation}
\label{eq:objective}
\begin{aligned}
\mathcal{L}=\ &\underbrace{\mathcal{L}^{w}_{\mathrm{pix}}
+\omega_{\mathrm{g}}\mathcal{L}^{w}_{\mathrm{grad}}
+\omega_{\mathrm{c}}\mathcal{L}_{\mathrm{coarse}}
+\omega_{\mathrm{p}}\mathcal{L}_{\mathrm{perc}}}
_{\text{reconstruction: every method}}\\[2pt]
&+\underbrace{\omega_{\mathrm{b}}\mathcal{L}_{\mathrm{bal}}
+\mathbf{1}_{n\ge3500}\bigl(\omega_{\mathrm{i}}\mathcal{L}_{\mathrm{inc}}
+\omega_{\alpha}\mathcal{L}_{t\alpha}
+\omega_{Y}\mathcal{L}_{tY}\bigr)}
_{\text{dictionary and routing: ours}},
\end{aligned}
\end{equation}
with $\omega_{\mathrm{g}}{=}0.15$, $\omega_{\mathrm{c}}{=}0.8$,
$\omega_{\mathrm{p}}{=}0.025$, $\omega_{\mathrm{b}}{=}0.01$,
$\omega_{\mathrm{i}}{=}0.03$, $\omega_{\alpha}{=}0.08$ and
$\omega_{Y}{=}0.02$ throughout, grid-searched once on held-out validation
clips and then frozen for every task, ablation and baseline. Balance runs from
step 0, since atoms that die early never recover, while the step gate defers
specialization until atoms have formed.

\section{Experiments}

\subsection{Experimental Setup}

\noindent\textbf{Evaluation.}
Every number below is computed on the \dataset{} test clips at native
$3840\times2160$ with no resizing or cropping. Fidelity is PSNR, SSIM and
LPIPS \citep{lpips}; no-reference quality MUSIQ \citep{musiq} and MANIQA
\citep{maniqa}; temporal stability warping error \citep{lai2018blind} and
tLPIPS \citep{tecogan} against the benchmark's released flow; cost
parameters and per-frame 4K runtime in FP16.

\noindent\textbf{Baselines.}
We compare against representative methods and current state-of-the-art
models: the image models AirNet \citep{airnet}, TransWeather
\citep{transweather}, DRNet \citep{drnet} and R2R \citep{r2r} with the
UHD specialist UHD-Processor \citep{uhdprocessor}, and the video models
EDVR \citep{edvr}, BasicVSR++ \citep{basicvsrpp}, Shift-Net
\citep{shiftnet} and RVRT \citep{rvrt} with the all-in-one ViWS-Net
\citep{viwsnet} and AverNet \citep{avernet}. All run blind, and the ones that
cannot fit a 4K frame in one pass share a single tiled path whose measured
cost enters the tables. Per-model disclosures are in the appendix.

\noindent\textbf{Training protocol.}
Every model, ours included, trains under one recipe: 100 epochs of AdamW at a
peak learning rate of $2\times10^{-4}$ with cosine decay, five-frame windows,
256-pixel native crops beside the short-side-512 coarse view, FP16, EMA, and
the reconstruction half of Eq.~\ref{eq:objective}, on eight V100 GPUs with
the four tasks mixed at every optimizer step.

\subsection{Main Results on \dataset{}}

\begin{figure*}[t]
\centering
\includegraphics[width=\textwidth]{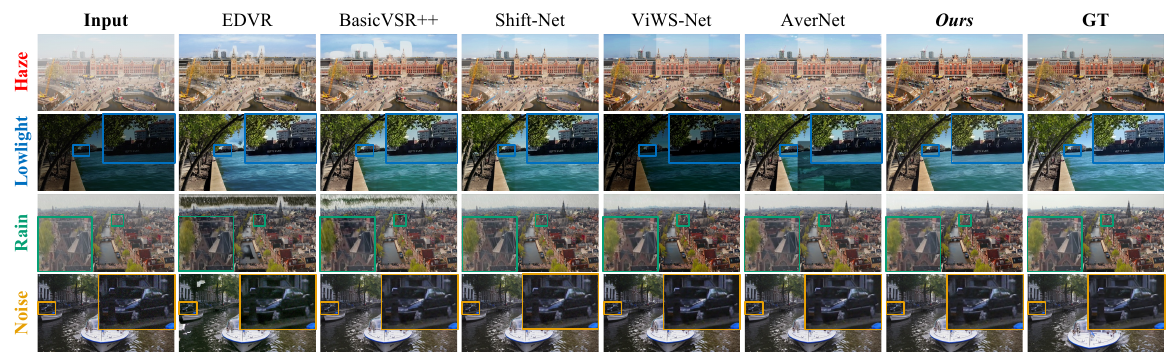}
\caption{\textbf{Restoration at native 4K}, one held-out clip per
degradation, all methods blind; boxes are 1:1 pixel crops. Each baseline
settles on its own white balance in the photometric rows, while \method{}
tracks the reference.}
\label{fig:qual}
\vspace{-10pt}
\end{figure*}

One set of weights covers the four degradations, and \method{} takes the best
task mean of Table~\ref{tab:main} while being the only method that restores a
4K frame in under a second: TransWeather, the fastest baseline, is 3.6 times
slower and AirNet, the most accurate one, 97 times.

The margin follows the split the model is built on. \method{} wins both
photometric tasks and beats every video method on rain, within 1.1\,dB of the
best image result. Denoising and the mean SSIM and LPIPS go to methods that
process every native pixel densely; that gap is band contention rather than
capacity, and the shallow refiner behind it is what keeps a 4K frame under half
a second. Figure~\ref{fig:qual} shows the rest: on rain and noise \method{}
clears the fine band without smoothing the structure behind it.

\noindent\textbf{Comparison with single-task experts.}
Unification is priced against experts of the same architecture and recipe, so
any gap is interference rather than content shift. The joint model gives up
1.04\,dB on average, and two-task subsets locate that loss, as
Table~\ref{tab:scope} records. Haze and low light
apply opposite corrections to global photometry and conflict as a pair, each
falling below its own expert; adding the two fine-band tasks overturns that and
lifts them 1.10 and 0.60\,dB above their experts. Rain and noise share a band
but not a direction: pairing denoising with deraining alone costs it 4.85\,dB,
nearly all of the 4.56\,dB it gives up in the full model, because one policy
suppresses unexplained detail everywhere and the other preserves everything but
streaks. That competition, not the photometric antagonism, is where the deficit
comes from.

\begin{table}[tb]
\centering
\caption{\textbf{Blind restoration under time-varying degradations.} The
degradation switches every $\Delta t$ frames inside each clip; Static repeats
the four-task mean of Table~\ref{tab:main}.}
\label{tab:tud}
\scriptsize
\setlength{\tabcolsep}{4pt}
\renewcommand{\arraystretch}{0.95}
\setlength{\aboverulesep}{0.35ex}
\setlength{\belowrulesep}{0.45ex}
\begin{tabular}{l|c|cc|cc}
\toprule
\multirow{2}{*}{Method} & \multirow{2}{*}{Static}
& \multicolumn{2}{c|}{$\Delta t{=}6$} & \multicolumn{2}{c}{$\Delta t{=}3$} \\
\cmidrule(lr){3-4}\cmidrule(l){5-6}
& & PSNR$\uparrow$ & SSIM$\uparrow$ & PSNR$\uparrow$ & SSIM$\uparrow$ \\
\midrule
AirNet & 27.11 & \underline{26.09} & \textbf{0.914} & 25.76 & \textbf{0.915} \\
TransWeather & 25.87 & 23.41 & 0.834 & 23.27 & 0.832 \\
DRNet & 18.62 & 23.17 & 0.820 & 23.05 & 0.824 \\
UHD-Processor & 26.77 & 24.05 & 0.746 & 24.19 & 0.752 \\
R2R & 15.75 & 26.02 & 0.829 & 25.59 & 0.828 \\
EDVR & 21.96 & 20.15 & 0.674 & 24.95 & 0.727 \\
BasicVSR++ & 26.25 & 23.87 & 0.864 & 25.81 & 0.885 \\
Shift-Net & 26.87 & 25.62 & 0.799 & \underline{27.14} & 0.821 \\
RVRT & 20.13 & 19.49 & 0.790 & 21.86 & 0.809 \\
ViWS-Net & 20.71 & 21.52 & 0.812 & 22.28 & 0.832 \\
AverNet & 26.92 & 22.36 & 0.714 & 24.94 & 0.756 \\
\midrule
\rowcolor{oursgray}\textbf{\method{} (Ours)} & 27.23 & \textbf{26.89} & \underline{0.898} & \textbf{27.16} & \underline{0.901} \\
\bottomrule
\end{tabular}
\vspace{-6pt}
\end{table}

\noindent\textbf{Robustness to TUD.}
Degradation in deployment does not hold still within a shot, so we re-read the
benchmark with the corruption switching inside each clip, a controlled variant
of the time-varying setting of \citet{avernet}. Both intervals reuse the
benchmark's own renders and present the same degradations in the same
proportions; only the speed differs.

A per-frame model cannot see the switch, and Table~\ref{tab:tud} shows it: no
image method moves more than half a decibel between the two intervals.
Every video baseline instead scores lower under slow switching than fast,
although the faster schedule is nominally the stress case, and AverNet, built
for this setting, loses 2.6\,dB. That is the cost of trusting neighbors: long
uniform segments reward the cross-frame fusion a switch then breaks, while at
$\Delta t{=}3$ the models learn to discount neighbors and fall back toward
per-frame behavior. \method{} gives up 0.3\,dB against its own static
score, because its degradation judgment is per-frame routing rather than
cross-frame alignment: a switch re-mixes atoms for one frame instead of
breaking a fused window, a response the appendix traces frame by frame.

\begin{table}[tb]
\centering
\caption{\textbf{Temporal stability} over the four tasks, one reference flow
for every method; the last two columns drop one temporal term.}
\label{tab:temp}
\small
\setlength{\tabcolsep}{2.8pt}
\resizebox{\columnwidth}{!}{%
\begin{tabular}{@{}l|cccc|ccc@{}}
\toprule
 & AirNet
 & \begin{tabular}[t]{@{}c@{}}Trans\\Weather\end{tabular}
 & \begin{tabular}[t]{@{}c@{}}UHD-\\Processor\end{tabular}
 & AverNet
 & \textbf{Ours}
 & \begin{tabular}[t]{@{}c@{}}w/o\\$\mathcal{L}_{t\alpha}$\end{tabular}
 & \begin{tabular}[t]{@{}c@{}}w/o\\$\mathcal{L}_{tY}$\end{tabular} \\
\midrule
Warping Error$\downarrow$
  & 0.0305 & 0.0329 & 0.0327 & 0.0296
  & \cellcolor{oursgray}0.0300 & 0.0319 & 0.0320 \\
tLPIPS$\downarrow$
  & 0.1256 & 0.1420 & 0.1551 & 0.1238
  & \cellcolor{oursgray}0.1375 & 0.1423 & 0.1434 \\
\bottomrule
\end{tabular}
}
\vspace{-6pt}
\end{table}

\begin{figure}[tb]
\centering
\includegraphics[width=0.9\linewidth]{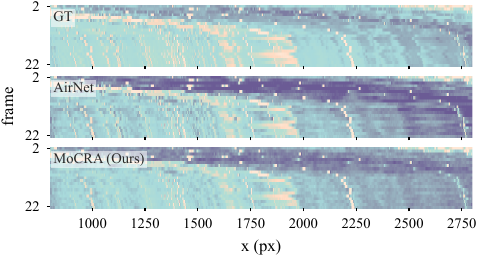}
\vspace{-6pt}
\caption{\textbf{Temporal stability on one scanline.} A fixed band of a
low-light test clip, shared luminance scale; the scene itself dims briefly
early on.}

\vspace{-10pt}
\label{fig:flicker}
\end{figure}

\noindent\textbf{Temporal stability.}
Table~\ref{tab:temp} keeps a representative subset; in the full roster of the
appendix, RVRT and EDVR, the two most invested in explicit alignment, post
the worst warping errors of the video group. Two methods there beat our warping
error and both charge for it: ViWS-Net reaches 0.0272 through a low-light
collapse whose near-black output has little left to flicker, and AverNet
reaches 0.0296 at 108 times our runtime per frame. Steadiness is not
smoothness in disguise when it arrives with the best task mean and no flow at
inference. Figure~\ref{fig:flicker} shows why: AirNet's per-frame
judgment follows every luminance change, including the ones it should ride
through, and the appendix's per-frame trace puts its steps a fifth larger
than ours. Dropping either temporal term returns a milder form of the
same flicker.

\begin{figure}[tb]
\centering
\begin{tikzpicture}[inner sep=0]
\node (lat) {\includegraphics{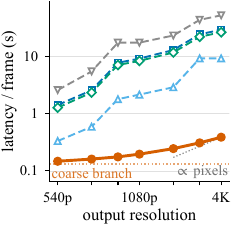}};
\node[anchor=south west] (mem) at ([xshift=4pt]lat.south east)
  {\includegraphics{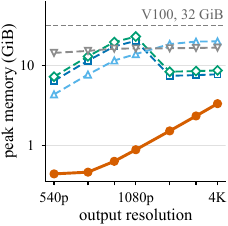}};
\node[anchor=north] at ([yshift=-0.8mm]lat.south east)
  {\includegraphics{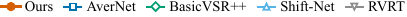}};
\end{tikzpicture}
\caption{\textbf{Scaling with output resolution} on one V100; solid is
a whole-frame pass, dashed the shared tiled path.}
\vspace{-2pt}
\label{fig:scale}
\end{figure}

\begin{figure}[tb]
\centering
\includegraphics[width=0.55\linewidth]{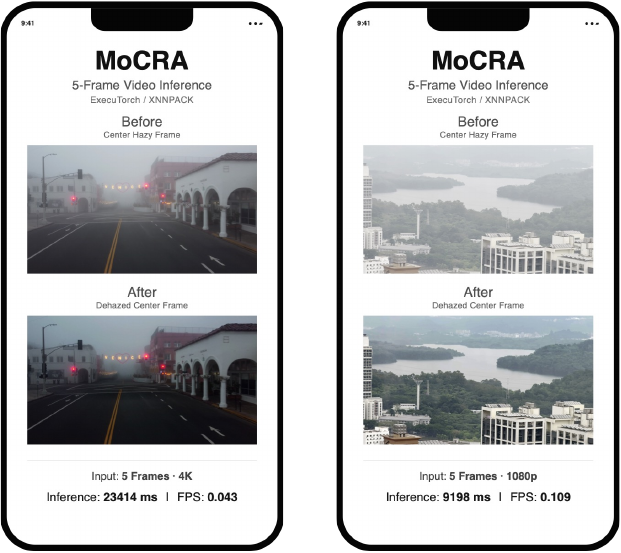}
\vspace{-2pt}
\caption{Phone demos on real degraded videos. Left: 4K street haze; right:
1080p scenery haze.}
\vspace{-10pt}
\label{fig:real}
\end{figure}

\noindent\textbf{Efficiency at native 4K.}
End-to-end, \method{} needs 0.48\,s and 3.9\,GiB per 4K frame on one GPU,
whole-frame; tiled baselines are timed as they run.
Figure~\ref{fig:scale}: globals are computed once per clip at fixed resolution,
so 1080p$\to$4K costs about $2\times$ latency and $3.8\times$ memory, with no
tiled fallback on a 32\,GB card; longer videos slide the window instead of
growing memory. Baselines hold 8--20\,GiB over the same sweep only by
shrinking tiles, paying in latency. The same checkpoint runs unchanged on a
phone via ExecuTorch/XNNPACK on real outdoor footage
(Figure~\ref{fig:real}).

\subsection{Ablations and Generalization}

\begin{table}[tb]
\centering
\caption{\textbf{Ablations on \dataset{}.} Per-task columns report PSNR; Mean
reports PSNR\,/\,SSIM.}
\label{tab:abl}
\scriptsize
\setlength{\tabcolsep}{1.5pt}
\resizebox{\columnwidth}{!}{%
\begin{tabular}{l|c|cccc|c}
\toprule
Variant & \shortstack{Params\\(M)}
& Dehaze & Derain & Denoise & Low light & Mean \\
\midrule
\rowcolor{oursgray}\textbf{\method{} (full)} & 3.64 & \textbf{23.86} & \textbf{28.71} & \textbf{32.33} & \textbf{24.83} & \textbf{27.43\,/\,0.900} \\
\midrule
w/o refiner atom sites & 3.34 & 21.74 & 26.79 & 30.37 & 23.23 & 25.53\,/\,0.878 \\
w/o cross-scale guidance & 3.62 & 22.29 & 27.24 & 31.38 & 23.65 & 26.14\,/\,0.889 \\
global temporal readout & 3.64 & 21.99 & 27.02 & 30.75 & 23.17 & 25.73\,/\,0.881 \\
\midrule
$+$ prototype memory & 3.85 & 21.79 & 27.50 & 30.99 & 23.41 & 25.92\,/\,0.882 \\
rank-2 atoms (matched) & 3.18 & 21.77 & 27.49 & 30.73 & 23.52 & 25.88\,/\,0.878 \\
\midrule
uniform fine-band loss & 3.64 & 22.29 & 27.30 & 30.79 & 23.07 & 25.86\,/\,0.878 \\
\bottomrule
\end{tabular}
}

\end{table}

Every variant in Table~\ref{tab:abl} retrains at the full budget and loses
1.3--1.9\,dB. The largest loss comes from taking the twelve atom sites out of
the refiner, and it spreads over all four tasks rather than concentrating in
the refiner's own band, which is what one shared dictionary predicts. A global
rather than per-position temporal readout gives up 1.7\,dB at identical size.
Shape matters more than size: rank-2 atoms at a matched budget fall 1.6\,dB
below rank-1, and a cross-clip prototype memory costs 1.5\,dB despite adding
capacity, by pulling routing toward its prototypes. Uniform supervision costs
1.6\,dB at unchanged size and runtime.

\begin{figure}[tb]
\centering
\scalebox{0.85}{%
\begin{tikzpicture}[inner sep=0,
  lab/.style={font=\footnotesize, inner sep=1pt}]
\node (tsne) {\includegraphics{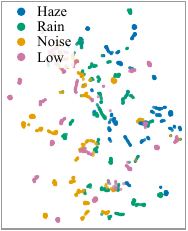}};
\node[anchor=south west] (usage) at ([xshift=4pt]tsne.south east)
  {\includegraphics{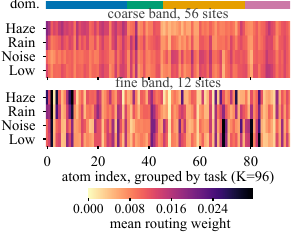}};
\node[lab, below=0.6mm of tsne.south] {(a)~Routing t-SNE};
\node[lab, below=0.6mm of usage.south] {(b)~Atom usage by band};
\end{tikzpicture}
}
\caption{\textbf{Routing anatomy on the \dataset{} test set.} (a) t-SNE of
per-window $\alpha$, clip means removed. (b) Task-mean weight per atom by band.}

\label{fig:routing}
\end{figure}

Figure~\ref{fig:routing} checks that the routing is not degenerate. It commits
without collapsing: per-frame entropy averages 2.23 against 2.48 for a uniform
twelve-atom mixture, and no atom of the 96 falls out of use. What it separates
is the band, not the task. Scene content dominates the raw routing, $\eta^2$
0.38 against 0.10 for the degradation, and a held-out linear task readout
reaches only 0.49 against 0.25 chance, while fine-band marginals peak at
3.1--4.7 times their own mean against 1.5--1.9 in the coarse band, the same
conditioning whose removal costs 1.9\,dB. Cross-task similarity is in the
appendix.

\noindent\textbf{Results on public benchmarks.}
We retrain the same setting on RainMotion \citep{rddnet} and REVIDE
\citep{revide} as one blind set outside our synthesis
(Table~\ref{tab:cross}). On
REVIDE's real haze, near the scale \method{} targets, it leads the roster
in PSNR and SSIM, carrying the Table~\ref{tab:main} margin onto footage we
did not render.

\section{Conclusion, Limitations and Future Works}

Restoring 4K video does not require giving up a single label-free model,
provided each degradation is matched to its own band: \method{} handles all
four blind at native resolution in 3.6M parameters and half a second per
frame, without optical flow, and exports unchanged to a phone runtime. Three
boundaries follow. Below the coarse view's 512 the two bands coincide and the
margin goes, joint training still owes the denoising expert four decibels, and
the phone path is a feasibility check rather than a real-time budget.
Quantizing for that budget, and composing unseen compound degradations from
simple atoms, come next.

\bibliography{aaai2027}

\clearpage
\raggedbottom
\setcounter{secnumdepth}{2}
\renewcommand{\thesection}{A\arabic{section}}
\renewcommand{\thesubsection}{\thesection.\arabic{subsection}}
\renewcommand{\thetable}{A\arabic{table}}
\renewcommand{\thefigure}{A\arabic{figure}}
\renewcommand{\theequation}{A\arabic{equation}}
\setcounter{section}{0}
\setcounter{subsection}{0}
\setcounter{table}{0}
\setcounter{figure}{0}
\setcounter{equation}{0}

\begin{center}
{\Large \textbf{Appendix}}
\end{center}
\vspace{1em}

\section{Benchmark Comparison}
Table~\ref{tab:datasets-full} situates \dataset{} among paired restoration
benchmarks.

\begin{table*}[t]
\centering
\caption{Where \dataset{} sits among paired restoration benchmarks: it is
the only one pairing four degradations as 4K video over a single shared
ground truth. \emph{Scale} gives train/test images, or clips$\times$frames
for video, with \emph{pooled}/\emph{stitched} marking assembly from a
different source per task; \emph{Aux.} lists released depth (d),
transmission (t), flow (f), noise level ($\sigma$), and illumination (i)
maps.}
\label{tab:datasets-full}
\small
\setlength{\tabcolsep}{6.5pt}
\begin{tabular}{ll|ccccccc}
\toprule
Dataset & Venue & Tasks & Type & Res. & Scale & Shared GT & Temporal & Aux. \\
\midrule
AirNet \citep{airnet}          & CVPR'22    & 3    & Image & $\le$512p   & pooled     & $\times$    & n/a      & $\times$ \\
PromptIR \citep{promptir}      & NeurIPS'23 & 3--5 & Image & $\le$512p   & pooled     & $\times$    & n/a      & $\times$ \\
CDD-11 \citep{onerestore}      & ECCV'24    & 11   & Image & $\le$512p   & 13k/2.2k   & $\checkmark$ & n/a     & $\times$ \\
4KID \citep{dehaze4kid}        & CVPR'21    & 1    & Image & 4K          & 8k/200     & n/a         & n/a      & $\times$ \\
4K-Rain13k \citep{derain4krain}& TMM'26     & 1    & Image & 4K          & 12.5k/500  & n/a         & n/a      & $\times$ \\
UHD-LL \citep{uhdll}           & ICLR'23    & 1    & Image & 4K          & 2k/150     & n/a         & n/a      & $\times$ \\
UHD-LOL4K \citep{uhdlol}       & AAAI'23    & 1    & Image & 4K          & 6k/2.1k    & n/a         & n/a      & $\times$ \\
UniFlowRestore \citep{uniflowrestore} & MM'25 & 4   & Video & $\le$1080p  & stitched   & $\times$    & part.    & $\times$ \\
UHV-4K \citep{libranet}        & arXiv'26   & 1    & Video & 4K          & 100$\times$25 & n/a      & $\checkmark$ & d/t/f \\
\midrule
\rowcolor{oursgray}\dataset{} (Ours) & --   & \textbf{4} & \textbf{Video} & \textbf{4K} & 100$\times$25 & $\checkmark$ & $\checkmark$ & d/t/f/$\sigma$/i \\
\bottomrule
\end{tabular}
\end{table*}

\section{Synthesis Implementation Details}

\subsection{Shared Geometry Pass}
The first pass runs once per clip. It estimates per-frame depth with
Depth-Anything-V2 (ViT-L) \citep{depthanythingv2} and forward flow with
RAFT \citep{raft} (Sintel weights, 20 iterations) on an aspect-preserving
960p proxy and caches both; the rain, noise, and low-light branches read
the haze branch's cache, so the geometry and motion behind all four
degradations are identical per frame. Disparity is flipped in polarity
and normalized by its per-clip 1st/99th percentiles onto $[0,d_{\max}]$
with $d_{\max}{=}10$. Low-frequency fields, depth, transmission, and the
rain veil, are upsampled to 4K with a guided filter \citep{6319316}
(radius 16, regularization $10^{-4}$), and transmission is clipped to
$[0.05,1]$.

\subsection{Severity Tiers}
Table~\ref{tab:tiers-full} lists the per-clip parameter tiers referred to in
the main text, together with the banding, channel mismatch, and color-cast
draws.

\begin{table}[t]
\centering
\caption{Per-clip degradation parameters by severity tier; each clip
freezes one setting and the 100 clips are tier-balanced. Noise terms are
linear-domain variance ($\sigma_{\mathrm{s}}^{2},\sigma_{\mathrm{r}}^{2}$)
or standard-deviation ($\sigma$)
scales. The low-light brightness guard forces near-black clips to the
mildest tier and demotes heavy assignments below median luminance 0.15,
shifting its tier balance to 37/42/21 while the other tasks stay at
34/33/33; color casts remain balanced at 34/33/33.}
\label{tab:tiers-full}
\footnotesize
\setlength{\tabcolsep}{4.5pt}
\begin{tabular}{l|ccc}
\toprule
Parameter & Light/Mild & Medium & Heavy \\
\midrule
\rowcolor{bandgray}\multicolumn{4}{c}{\emph{Haze}} \\
\midrule
scattering $\beta$ & 0.04--0.08 & 0.08--0.15 & 0.15--0.30 \\
atmos.\ light $A$ & \multicolumn{3}{c}{$\{0.85,0.90,0.95\}$ per clip} \\
\midrule
\rowcolor{bandgray}\multicolumn{4}{c}{\emph{Rain}} \\
\midrule
streak density (/Mpx) & 180 & 540 & 1080 \\
fall speed (px/frame) & 20 & 40 & 70 \\
streak opacity & 0.20--0.40 & 0.30--0.60 & 0.55--0.85 \\
veil $\beta_r$ & 0.008 & 0.020 & 0.040 \\
veil light $A_r$ & \multicolumn{3}{c}{$\{0.55,0.65,0.75\}$ per clip} \\
dominant angle $\theta$ & \multicolumn{3}{c}{$[-30^\circ,+30^\circ]$ per clip} \\
\midrule
\rowcolor{bandgray}\multicolumn{4}{c}{\emph{Noise}} \\
\midrule
shot $\sigma_{\mathrm{s}}^{2}$ ($\times10^{-3}$) & 0.9--2.0 & 3.5--8.0 & 13.5--25.0 \\
read $\sigma_{\mathrm{r}}^{2}$ ($\times10^{-5}$) & 0.1--0.5 & 1.0--5.0 & 12--40 \\
FPN $\sigma$ ($\times10^{-3}$) & 0.3--0.7 & 0.8--1.5 & 2.0--3.5 \\
band $\sigma$ ($\times10^{-4}$) & 1.5--3.0 & 3.0--5.5 & 6--11 \\
channel mismatch & $\pm10\%$ & $\pm12\%$ & $\pm15\%$ \\
equivalent ISO & ${\sim}1600$ & ${\sim}6400$ & ${\sim}25600$ \\
\midrule
\rowcolor{bandgray}\multicolumn{4}{c}{\emph{Low light}} \\
\midrule
exposure $a$ & 0.20--0.30 & 0.10--0.20 & 0.05--0.10 \\
tone $\gamma$ & 1.2--1.6 & 1.6--2.4 & 2.4--3.2 \\
vignette $k$ & 0.00--0.10 & 0.10--0.20 & 0.20--0.32 \\
color cast $\mathbf{g}$ & \multicolumn{3}{c}{\{warm, neutral, cool\} per clip} \\
\bottomrule
\end{tabular}
\end{table}

\subsection{Branch Models and Constants}
\noindent\textbf{Haze.}
The branch is adopted unchanged from UHV-4K \citep{libranet}: the
atmospheric scattering model
\begin{equation}
I = J\,\mathcal{T} + A\,(1-\mathcal{T}), \qquad \mathcal{T} = \exp(-\beta\,d),
\end{equation}
with the clean frame $J$, normalized depth $d$, and clip-level scattering
coefficient $\beta$ and atmospheric light $A$, the latter assigned
independently of the $\beta$ tier. Depth is smoothed along the flow,
$d_t \leftarrow 0.8\,d_t + 0.2\,\mathrm{warp}(d_{t-1})$, before the
transmission is formed, which keeps the haze stable under camera motion.

\noindent\textbf{Rain.}
The veil is the same scattering model at a much weaker coefficient,
\begin{equation}
I_{\text{veil}} = J\,\mathcal{T}_r + A_r\,(1-\mathcal{T}_r),
\qquad \mathcal{T}_r = \exp(-\beta_r\,d),
\end{equation}
with $\mathcal{T}_r$ floored at $0.5$ so distant scenery is never veiled
into haze.
The particle field seeds $\mathrm{density}\times(HW/10^{6})$ particles at
uniform subpixel positions on the proxy plane; each carries a $\pm15\%$
speed jitter, the clip's dominant angle jitters by $\pm3^\circ$ per
frame, opacity decays by $0.85$ per frame, and a particle respawns in a
100-pixel band at the top edge once it leaves the frame, exceeds a
12-frame lifetime, or fades out. Streaks are drawn at 4K as one-pixel
anti-aliased line segments whose length equals the per-frame
displacement, at intensity 0.95, then blurred with a Gaussian of
$\sigma=0.8$. Rasterizing at native scale matters: rendering the streaks
on the proxy and upsampling with the guided filter attenuates one-pixel
lines by 35--40\%.

\noindent\textbf{Noise.}
The shot variance $\sigma_{\mathrm{s}}^{2}$ is drawn per channel with the
tier's mismatch across the three, which mirrors the heterogeneity left by
demosaicing, and the per-clip draws are calibrated against raw sensor
measurements \citep{crvd}. The exposure gain follows an AR(1) process,
$g_t = 1+\delta_t$ with $\delta_t = 0.92\,\delta_{t-1} + e_t$ and
$\operatorname{std}(e_t) = 0.008$, a steady-state flicker of about
2\%. Banding adds one row and one column offset vector per frame at the
tier's band $\sigma$. The released noise-level map is the per-pixel
total standard deviation,
\begin{equation}
\sigma^{2}_{\text{tot}} = \sigma^{2}_{\mathrm{s},c} \max(g_t\,x, 0)
+ \sigma^{2}_{\mathrm{r}} + \sigma^{2}_{\text{fpn}}
+ 2\,\sigma^{2}_{\text{band}},
\end{equation}
the factor 2 counting the independent row and column draws.

\noindent\textbf{Low light.}
The cascade applies, in order, the exposure drop $a\,x^{\gamma}$, the
clip-level cast $\mathbf{g}$, the vignette
$V(r) = 1 - k\,(r/r_{\max})^{2}$ with $r$ the distance from the image
center, and the AR(1) gain $g_t$, before the ISO-tier sensor noise is
added. The cast is drawn from warm $(1.10, 1.00, 0.75)$, neutral
$(1,1,1)$, and cool $(0.78, 0.95, 1.18)$ nominals in RGB with $\pm4\%$,
$\pm2\%$, and $\pm4\%$ jitter; the nine darkness-by-cast combinations
are balanced over the 100 clips, and darkness tiers map one-to-one onto
the noise ISO tiers, so darker clips carry heavier noise. The brightness
guard triggers on real content: four night scenes with median luminance
below 0.05 are forced to the mildest tier, and heavy assignments below
0.15 are demoted to medium.

\subsection{Seeding and Caching}
All randomness is drawn from per-clip, per-component streams: the seed
for clip $c$ and component $s$ (fixed pattern, flicker, shot noise,
\dots) is $\mathrm{MD5}(c \,\Vert\, s \,\Vert\, b)$ reduced to 31 bits,
with base seed $b{=}42$ for haze, noise, and low light and $b{=}142$ for
rain, which decorrelates its draws from the haze parameters. The frozen
sensor patterns and the 25-frame gain sequences are cached per clip, so
the same code and configuration reproduce every released frame.

\subsection{Released Modalities and Layout}
Every modality ships per frame under a type-first layout, one directory
per modality (\texttt{gt}, the four degraded inputs, \texttt{depth},
\texttt{transmission}, \texttt{flow}, \texttt{noise\_sigma\_map},
\texttt{illumination\_map}) holding one subdirectory per clip;
Table~\ref{tab:modalities} lists resolutions and encodings. The flow is
released at its native 960p proxy resolution and is shared by all four
tasks. The 20 test clips are identical for every task: 00005, 00008,
00009, 00018, 00023, 00024, 00029, 00032, 00040, 00042, 00046, 00047,
00058, 00069, 00074, 00079, 00082, 00084, 00089, 00099.

\begin{table}[t]
\centering
\caption{Released modalities. All images are 8-bit PNG; the flow is
float32 per-pixel displacement at the 960p proxy resolution.}
\label{tab:modalities}
\small
\setlength{\tabcolsep}{4pt}
\begin{tabular}{lll}
\toprule
Modality & Res. & Encoding \\
\midrule
GT, degraded inputs & 4K & PNG \\
depth & 4K & $\mathrm{clip}(d/d_{\max}\cdot 255)$ \\
transmission & 4K & $\mathcal{T}\cdot 255$ \\
flow & 960p & \texttt{.npy}, $[H,W,2]$ \\
noise level & 4K & $\mathrm{clip}(\sigma_{\text{tot}}\cdot 255)$ \\
illumination & 4K & $a\,\mathbf{g}\,V(r)\cdot 255$ \\
\bottomrule
\end{tabular}
\end{table}

\section{Consistency Verification Protocol}
Each branch ships a quality-check script that inverts the synthesis on
the released frames and compares the estimates against the per-clip
assignment records; a clip passes only if every check holds. All four
branches share modality-alignment and pixel-range checks; the
branch-specific inversions follow.

\noindent\textbf{Dehaze.}
The atmospheric light re-estimated from far-field radiance must satisfy
$|\hat{A}-A|\le 0.03$; the 95th percentile of the depth change between
adjacent frames must stay below $0.05$, and the mean frame-to-frame
radiance change below $15$ on the 8-bit scale.

\noindent\textbf{Derain.}
The previous frame's streak mask, warped by the released flow, must
overlap the current mask with IoU in $[0.10, 0.85]$, which rules out
both frozen and uncorrelated rain; the streak angle must hold its clip-level
value within a drift of $\sigma\le 10^\circ$, streak sparsity must match
the assigned tier, and the tier-by-veil assignment table must remain
balanced.

\noindent\textbf{Denoise.}
A binned fit of
$\mathrm{Var} = \sigma_{\mathrm{s}}^{2} x + \sigma_{\mathrm{r}}^{2}$ to the
residuals must
recover the assigned noise law; the mean residual across frames must
reproduce the frozen sensor pattern within $\pm50\%$ of its $\sigma$;
and the per-frame noise residual must exceed the ground-truth warping
residual, which confirms that the noise does not follow the scene.

\noindent\textbf{Low light.}
The output must be darker than the ground truth with a consistent
frame-to-frame brightness ratio; a log--log fit must recover the tone
curve within $|\hat{\gamma}-\gamma| < 1$; and the corner-to-center
luminance ratio must reproduce the assigned vignette, $\approx 1-k$.

\section{Inference Procedure at 4K}
A clip is first reflect-padded to a multiple of the refiner's unshuffle
factor, so that every window start stays exactly representable on the
half-resolution grid the guidance lives on; the padding is cropped off at the
end and is a no-op at benchmark sizes. The whole frame is then reduced to the
short-side-512 coarse view with the antialiased bilinear kernel the training
loader uses, and the Atom Composer reads a further 256-pixel view of it.
Routing tensor, coarse output and guidance latent each come from one pass over
that view.

The coarse output is upsampled to native resolution and the refiner runs over
native windows, the three global objects sliced rather than recomputed under
the coordinate mapping training crops use. Whole-frame and tiled execution
share this window code, a single window being the case whose cosine taper
cancels exactly. The whole frame goes in one pass whenever a conservative
estimate of the refiner's peak activations stays under half the free device
memory; only a genuine out-of-memory error falls back to the overlapping tile
grid, which blends with the same taper and runs its windows in batches. Every
other exception propagates, so a shape or logic error cannot be masked as a
memory event.

\section{Training Objective Details}
Two terms of the main text's objective guard the dictionary.
$\mathcal{L}_{\mathrm{inc}}$ penalizes the mean absolute cosine between the
output factors of each site's most overlapping atom pairs, which keeps atoms
from converging onto one another.
$\mathcal{L}_{\mathrm{bal}}$ is the KL divergence from uniform to the global
atom-usage marginal, taken in the direction that penalizes dead atoms without
bound; the marginal is reduced across ranks, so a per-rank single-task batch
cannot force every task to use the dictionary uniformly.

The temporal term $\mathcal{L}_{tY}$ on the coarse output normalizes over all
positions rather than over the gate's support, which lets it vanish under
global motion instead of penalizing it.

\section{Baseline Training and Inference Disclosures}
Every baseline trains under the recipe of the main text; departures are
memory-driven and were granted in a fixed order: gradient checkpointing
and an expanding allocator first, both of which leave the computation
unchanged; then batch 1 with two accumulation steps, which preserves the
effective batch and which we withhold from models whose batch statistics are
live; then an official smaller variant where the authors publish one; then the
method's own published crop. Among the eleven baselines of the main table,
Shift-Net and RVRT train at batch 1 with accumulation and EDVR is the official
EDVR-M variant. None needed its published crop, none was dropped, and no cell
of the main table reports an out-of-memory failure. The two flow-based video
baselines use public pretrained SpyNet weights, frozen.

At test time every baseline that cannot hold a 4K frame takes one shared tiled
path: tiles start at 1536 with an overlap of 32 and cosine-tapered blending,
and halve on a capacity failure until the model fits; tiled evaluation is
itself standard, the official test scripts of VRT, RVRT and SwinIR exposing
the same switch. The output is always native 4K,
never a downsampled pass upsampled back, so a smaller tile is charged to a
method's measured runtime rather than to its quality protocol. Three ports
needed numerical guards under FP16 that their authors, who train in FP32, never
exercised: EDVR clamps its deformable offsets and runs its forward pass in
FP32, RVRT keeps its attention kernel in FP32, and R2R fills its retrieval mask
with the smallest representable value instead of a constant that overflows half
precision. Each guard is bit-exact with the original in FP32.

\section{Single-Task Experts and Degradation Pairs}
Table~\ref{tab:scope} gives the per-task scores behind the expert comparison of
the main text. \emph{Single} trains one model per degradation, \emph{Pair} on
the two subsets that isolate the photometric and fine-band conflicts, and
\emph{All-4} is the joint model of Table~1 in the main text.

\begin{table}[t]
\centering
\caption{PSNR across training scopes, same architecture and recipe throughout.}
\label{tab:scope}
\small
\setlength{\tabcolsep}{8pt}
\begin{tabular}{@{}lcccc@{}}
\toprule
Task & Single & Pair & All-4 & $\Delta$ \\
\midrule
Dehaze & 22.37 & 21.82 & 23.47 & $+1.10$ \\
Low light & 24.06 & 23.89 & 24.66 & $+0.60$ \\
\midrule
Derain & 30.03 & 30.08 & 28.70 & $-1.33$ \\
Denoise & 36.63 & 31.78 & 32.07 & $-4.56$ \\
\midrule
Mean & 28.27 & 26.90 & 27.23 & $-1.04$ \\
\bottomrule
\end{tabular}
\end{table}

\section{Temporal Stability: Full Roster}
Table~\ref{tab:temp-full} gives all eleven baselines for the temporal
metrics of the main text, which reports a representative subset. Explicit
alignment does not buy stability on degraded footage: RVRT and EDVR, the two
restorers most invested in it, post the two worst warping errors of the video
group. ViWS-Net's leading scores come with a low-light collapse (8.44\,dB in
the main comparison, a near-black output that has little left to flicker),
so the calmest honest row of the roster is AverNet, at 108 times \method{}'s
per-frame runtime.

\begin{table}[t]
\centering
\caption{Temporal stability averaged over the four tasks, computed against
the same reference flow for every method. Bold and underline mark best and
second best; the last two rows each remove one temporal term from
\method{}.}
\label{tab:temp-full}
\small
\setlength{\tabcolsep}{14.5pt}
\begin{tabular}{l|cc}
\toprule
Method & Warping Error$\downarrow$ & tLPIPS$\downarrow$ \\
\midrule
\rowcolor{bandgray}\multicolumn{3}{c}{\emph{(a) Image methods}} \\
\midrule
AirNet & 0.0305 & 0.1256 \\
TransWeather & 0.0329 & 0.1420 \\
DRNet & 0.0301 & 0.1428 \\
UHD-Processor & 0.0327 & 0.1551 \\
R2R & 0.0494 & 0.1601 \\
\midrule
\rowcolor{bandgray}\multicolumn{3}{c}{\emph{(b) Video methods}} \\
\midrule
EDVR & 0.0427 & 0.1566 \\
BasicVSR++ & 0.0327 & 0.1403 \\
Shift-Net & 0.0320 & 0.1495 \\
RVRT & 0.0441 & 0.2050 \\
ViWS-Net & \textbf{0.0272} & \textbf{0.1237} \\
AverNet & \underline{0.0296} & \underline{0.1238} \\
\midrule
\rowcolor{oursgray}\textbf{\method{} (Ours)} & 0.0300 & 0.1375 \\
\quad w/o $\mathcal{L}_{t\alpha}$ & 0.0319 & 0.1423 \\
\quad w/o $\mathcal{L}_{tY}$ & 0.0320 & 0.1434 \\
\bottomrule
\end{tabular}
\end{table}

\begin{figure}[t]
\centering
\includegraphics{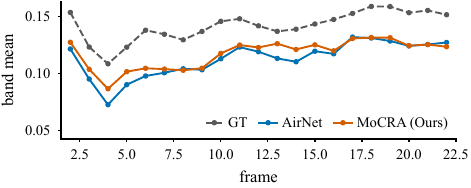}
\caption{Mean intensity per frame of the scanline band shown in the main
text, on the same low-light test clip.}
\label{fig:flicker-trace}
\end{figure}

Figure~\ref{fig:flicker-trace} carries the per-frame band means behind the
scanline figure of the main text. The mean frame-to-frame step is 0.0076 for
AirNet against 0.0062 for \method{}, and the band mean swings by 0.059 against
0.045, with the reference clip at 0.050.

\section{Cross-Task Routing Similarity}
\begin{figure}[t]
\centering
\includegraphics{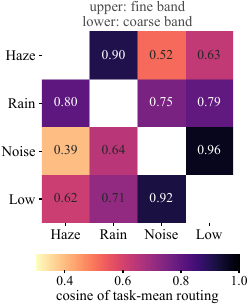}
\caption{Cosine between task-mean routing vectors on the standard four-task
test set, computed per band from the same routing collection as the
main-text figure.}
\label{fig:routing-cosine}
\end{figure}

Figure~\ref{fig:routing-cosine} completes the routing anatomy of the main
text with the similarity structure across tasks, and it rules out a
dictionary silently partitioned into four disjoint experts: no pair of
tasks routes orthogonally. The ordering of the pairs instead reads the
benchmark's physics back. Noise and low light sit highest, 0.92 in the
coarse band and 0.96 in the fine one, and their renders share one sensor
model; haze and rain follow at 0.80 and 0.90, the rain branch compositing
its streaks over a thin veil drawn from the same scattering model; haze and
noise sit lowest at 0.39 and 0.52, sharing nothing but the clean frame.

\section{Time-Varying Degradations: Per-Frame Analysis}
\begin{figure}[t]
\centering
\includegraphics{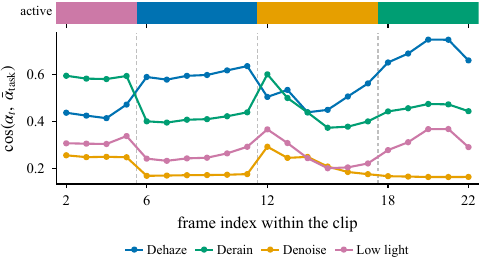}
\caption{Per-frame routing under the switching schedule ($\Delta t{=}6$). Each
curve is the cosine similarity between the frame's routing vector and one
task's mean routing profile; the strip above shows the active degradation,
and dashed lines mark the switches. The model is retrained on the switching
data and receives no label at any point.}
\label{fig:tud-routing}
\end{figure}

Figure~\ref{fig:tud-routing} traces how the routing of the
$\Delta t{=}6$-retrained model moves through one held-out switching clip. Each
curve is the cosine between the frame's routing vector $\alpha_t$ and one
task's mean profile $\bar{\alpha}_{\text{task}}$, the profiles collected
from the same checkpoint reading the standard four-task test set. The
schedule is legible as movement rather than level: over the 60 switches of
the 20 test clips, similarity to the newly active degradation rises by
0.10 at the switch frame and similarity to the outgoing one falls by 0.11,
against a mean per-frame drift of 0.05 away from the boundaries; three
quarters of all switches move toward the incoming task, 83\% move away
from the outgoing one. The absolute height of each curve is set by the
clip's content and by the overlap among the four mean profiles, whose
pairwise cosines Figure~\ref{fig:routing-cosine} puts between 0.39 and
0.96, so the active task's curve need not ride highest. What the
trace establishes is the response granularity: the routing re-mixes within
one frame of each switch, which is the mechanism behind the robustness
table of the main text.

\section{Results on Public Benchmarks}
Table~\ref{tab:cross} gives the full roster and resolution detail deferred
from the external-benchmark paragraph of the main text. RainMotion
\citep{rddnet} contributes rain from an unrelated synthesis model over
unrelated scenes at $640\times480$, REVIDE \citep{revide} real indoor haze
registered against clean passes at $2708\times1800$. Merged, they put real
capture into the training set, and every method retrains on them under the
recipe of the main text.

The split along resolution reads the method's own premise back to it. On
REVIDE, whose frames approach the scale \method{} is built for, it posts the
best PSNR and SSIM of the roster on real haze. On RainMotion the ranking
inverts and \method{} falls 10\,dB behind the flow-based propagation of
AverNet and BasicVSR++, which is cheap and effective at $640\times480$. That
is the design premise withdrawn rather than a training accident: the frame's
short side sits below the coarse view's 512, so the two scales the
architecture separates coincide and what remains is a 3.6M-parameter
single-scale model against full-size restorers.

\begin{table}[t]
\centering
\caption{External public benchmarks. Joint blind training on RainMotion
\citep{rddnet} and REVIDE \citep{revide}; each benchmark is scored at its own
native resolution. Bold and underline mark best and second best.}
\label{tab:cross}
\scriptsize
\setlength{\tabcolsep}{6pt}
\renewcommand{\arraystretch}{0.95}
\setlength{\aboverulesep}{0.35ex}
\setlength{\belowrulesep}{0.45ex}
\begin{tabular}{l|cc|cc}
\toprule
\multirow{2}{*}{Method} & \multicolumn{2}{c|}{RainMotion (derain)}
& \multicolumn{2}{c}{REVIDE (dehaze)} \\
\cmidrule(lr){2-3}\cmidrule(l){4-5}
& PSNR$\uparrow$ & SSIM$\uparrow$ & PSNR$\uparrow$ & SSIM$\uparrow$ \\
\midrule
\rowcolor{bandgray}\multicolumn{5}{c}{\emph{(a) Image methods}} \\
\midrule
AirNet & 28.43 & \underline{0.917} & 19.11 & 0.845 \\
TransWeather & 22.76 & 0.830 & 19.52 & 0.848 \\
DRNet & 16.55 & 0.539 & 19.11 & 0.823 \\
UHD-Processor & 25.41 & 0.816 & 18.95 & 0.834 \\
R2R & 26.88 & 0.890 & \underline{20.31} & \underline{0.855} \\
\midrule
\rowcolor{bandgray}\multicolumn{5}{c}{\emph{(b) Video methods}} \\
\midrule
EDVR & 28.96 & 0.916 & 19.68 & 0.851 \\
BasicVSR++ & \underline{31.53} & \textbf{0.935} & 18.42 & 0.844 \\
Shift-Net & 27.95 & 0.893 & 19.43 & 0.841 \\
RVRT & 24.30 & 0.812 & 17.36 & 0.767 \\
ViWS-Net & 16.68 & 0.567 & 17.05 & 0.807 \\
AverNet & \textbf{31.77} & \textbf{0.935} & 18.84 & 0.845 \\
\midrule
\rowcolor{oursgray}\textbf{\method{} (Ours)} & 21.79 & 0.800 & \textbf{21.16} & \textbf{0.861} \\
\bottomrule
\end{tabular}
\end{table}

\end{document}